# Notes on the H-measure of classifier performance

**David J. Hand and Christoforos Anagnostopoulos**
Department of Mathematics, Imperial College, London
{d.j.hand@imperial.ac.uk}

**Abstract:**
The H-measure is a classifier performance measure which takes into account the context of application without requiring a rigid value of relative misclassification costs to be set. Since its introduction in 2009 it has become widely adopted. This paper answers various queries which users have raised since its introduction, including questions about its interpretation, the choice of a weighting function, whether it is strictly proper, its coherence, and relates the measure to other work.

**1. Introduction**

Two-class classification is made on the basis that some objects have a certain property X while other objects do not. Most statistical and machine learning classification methods achieve this by initially estimating the probability that objects belong to each class. The probability estimates are based on observed properties of the objects, which one believes to be related to the presence/absence of X. Predicted classes are then obtained by comparing the estimated probabilities of belonging to one of the classes with a threshold, with objects with a value above the threshold being predicted to belong to that class, and other objects to the other class. The choice of the threshold will depend on the problem and the objectives. However, except in certain special applications, errors occur, with some objects ending up misclassified. In such cases, the choice of threshold must be made to balance the two types of error to yield an overall "best" performance. Since "balance" and "best" can be defined in various ways, this has led to many different ways of defining and measuring classifier performance (for reviews, see for example, Hand, 1997, 2012; Liu *et al*, 2014; Jiao and Du, 2016; Tharwat, 2020).

One measure of classifier performance is the H-measure (Hand, 2009, 2010; Hand and Anagnostopoulos, 2014). This paper presents a reformulated introduction to the H-measure, describing what has been learnt about its performance since its introduction, and answering questions which have been raised about it.

The H-measure can be approached in various ways. We originally approached it (Hand, 2009) as a consequence of identifying what is, from one perspective, a fundamental conceptual weakness in another widely used measure of classifier performance – the Area Under the Receiver Operating Characteristic curve (AUC). We later discovered that Buja *et al* (2005) had also derived it, in a more general context, approaching it from the direction of understanding boosting methods for improving statistical classification methods (Friedman,

Hastie, and Tibshirani, 2000). Although the body of this paper takes our perspective, in Section 8 we examine the broader description of Buja *et al* (2005).

Section 2 reprises the definition of the H-measure, its relationship to the AUC, and hence its motivation arising from the weakness of that measure. Section 3 describes the choice of the *weight* function, which is central to the definition of the H-measure. Section 4 gives a simple interpretation of the H-measure. Section 5 explores whether the H-measure is a so-called *strictly proper* scoring rule. Section 6 looks at how to handle unknown class sizes. Section 7 describes calculation of the H-measure. Section 8 then takes a step back and summarises the elegant derivation of the H-measure and related measures in a broader context given by Buja *et al* (2005). Section 9 looks at classification contexts where other ways of choosing the classification threshold are appropriate. Section 10 draws some conclusions.

## 2. Background

Given two classes of objects, labelled 0 and 1, our aim is to assign new objects to their correct class. To do this, we use the information in a sample of objects each with known class memberships (this is called the training set or design set) to construct an algorithm, a classification model, a classification rule, a measuring instrument, or a diagnostic system which will yield estimated probabilities of belonging to class 1 (and, by implication, also for class 0) for future objects. Let $q$ be the estimated probability that an object belongs to class 1, and define $F_0(q)$ and $F_1(q)$ to be the cumulative distribution functions of these estimates of probability for objects randomly drawn from classes 0 and 1 respectively, $\pi_0$ and $\pi_1$ to be the respective class sizes (alternative terms sometimes used for the latter are class proportions, ratios, or priors; we discuss how to handle uncertain class sizes in Section 6), $F(q) = \pi_0 F_0(q) + \pi_1 F_1(q)$ to be the overall mixture cumulative distribution function, and $f_0(q), f_1(q)$, and $f(q) = \pi_0 f_0(q) + \pi_1 f_1(q)$ to be the corresponding density functions. For a given training set and a specified way of deriving the classification model, these functions are known. This paper is concerned with estimating the performance of a classification rule based on such known functions. Comparison of performance estimates for different rules can be used to choose between rules, and to adjust or optimise a rule (e.g. enabling parameter optimisation, threshold setting, etc). However, note that this paper is not concerned with estimating or comparing the performance of methods of *constructing* classification or machine learning rules, which would require also taking into consideration the randomness implicit in the choice of training sample and random elements in the estimation algorithm (e.g. in the case of random forests or cross-validation).

To classify an object, its estimated probability of class 1 membership, or "score" for short, is compared with a threshold *t*, the "classification threshold", such that objects with scores greater than *t* are assigned to class 1 and otherwise to class 0. This results in two kinds of potential misclassifications: a class 0 object might be misclassified as class 1, and a class 1 object might be misclassified as class 0. Define $c \in [0,1]$ to be the cost due to misclassifying a class 0 object and $(1-c)$ the cost due to misclassifying a class 1 object, and take correct classifications as incurring no cost. This is the basic and most common situation encountered, but various generalisations can be made. For example, correct classifications might incur a cost (in that case the cost scale can be standardised by subtracting the cost of a correct classification and renormalising *c*); misclassification costs might differ depending on how "severe" is the misclassification (an object which has a score just on the "wrong" side of the threshold might incur a different cost from one which is far from the threshold on the wrong side); misclassification costs (and even the correct classification costs) might depend on the object, yielding a distribution of costs across the population; costs might not combine additively; population score distributions might change over time (sometimes called *population drift* or *concept drift*. This is often handled by revising the classifier periodically. For example, in credit scoring the tradition has been to rebuild every three years or so, but more advanced methods adaptively update the parameters); and, perhaps the most important generalisation, more than two classes might be involved (extending the H-measure to more than two classes is an ongoing project).

We see that, when threshold $t$ is used, the total cost due to misclassification is
$$L(c;t) = c\pi_0\big(1 - F_0(t)\big) + (1 - c)\pi_1 F_1(t). \qquad [1]$$
For any given cost $c$ a sensible choice of classification threshold $t$ is that which minimises $L(c;t)$ (we consider other ways of choosing $t$ in Section 9). Assuming the distributions are differentiable on $[0,1]$, differentiating [1] shows that this is given by $t = T_c$ satisfying
$$c = \pi_1 f_1(T_c)/f(T_c). \qquad [2]$$
But (by definition of $\pi_1$, $f_1(q)$, and $f(q)$) the ratio $\pi_1 f_1(q)/f(q)$ is the estimated conditional probability that an object with score $q$ will belong to class 1 – estimated from the training set using the specified classifier model or algorithm. That is $\pi_1 f_1(q)/f(q) = q$. It follows from [2] that the (estimated) best choice of threshold to use when $c$ is the cost of misclassifying a class 0 object and $(1 - c)$ is the cost of misclassifying a class 1 object is $T_c = c$.

This leads to a minimum classification loss of
$$L(c) = c\pi_0\big(1 - F_0(c)\big) + (1 - c)\pi_1 F_1(c).$$

In summary, if the cost $c$ (and class sizes and distributions) are known, this gives us the unique classification threshold to use and the associated consequent minimum loss. In general, however, costs are difficult to determine and they may not be known at the time that the classification rule has to be evaluated. For example, in a clinical setting the severity of misclassifications might depend on what treatments will be available in a clinic or on the particular characteristics of the local patient population, or in a credit scoring context the degree of acceptable risk might depend on current interest rates. More generally, we might want to make comparative statements about the relative performance of classification rules without knowing the details of the environment in which they will be applied, with the risk that any particular choice of costs could be very different from those encountered in practice. For these reasons, we will integrate over a distribution of costs, $w(c)$ say, chosen to represent one's beliefs about the costs to be encountered in the future. A similar point applies if the class sizes, $\pi_0$ and $\pi_1$, are unknown, but they can, at least in principle, be estimated from empirical considerations. We discuss this further in Section 6.

The overall expected minimum misclassification loss is then
$$L = \int_0^1 \big[c\pi_0\big(1 - F_0(c)\big) + (1 - c)\pi_1 F_1(c)\big]w(c)dc. \qquad [3]$$

Substituting $c = \pi_1 f_1(c)/f(c)$ into [3] yields

$$L = \int_0^1 \left[\frac{\pi_1 f_1(c)}{f(c)}\pi_0\bigl(1 - F_0(c)\bigr) + \frac{\pi_0 f_0(c)}{f(c)}\pi_1 F_1(c)\right] w(c)\,dc$$

$$= \pi_0\pi_1 \int_0^1 \bigl[f_1(c)\bigl(1 - F_0(c)\bigr) + f_0(c)F_1(c)\bigr]\frac{w(c)}{f(c)}\,dc$$

From this we see that, were we to take $w(c) = f(c)$, we would obtain $L = L_A$ where

$$L_A = \pi_0\pi_1 \int_0^1 \bigl[f_1(c)\bigl(1 - F_0(c)\bigr) + f_0(c)F_1(c)\bigr]\frac{f(c)}{f(c)}\,dc$$
$$= 2\pi_0\pi_1\left(1 - \int_0^1 F_0(c)f_1(c)\,dc\right) \qquad [4]$$

The expression $\int_0^1 F_0(c)f_1(c)\,dc$ in [4] is the familiar and widely used measure of performance known as the *Area Under the Receiver Operating Characteristic Curve*, denoted AUC (see, for example, Krzanowski and Hand, 2009).

Inverting [4], we obtain
$$AUC = 1 - L_A/2\pi_0\pi_1.$$

That is, the AUC for a particular classifier is a linear function of the expected misclassification loss when the cost distribution $w$ is taken to be the overall score distribution for the classifier being evaluated: $w(c) = f(c)$.

This seems to provide a justification for the widespread use of the AUC in comparing classification rules. However, note that the overall score distribution $f$ will generally be different for different classifiers. This means that, if we choose the distribution $w$ to be $f$, the loss $L_A$ will be calculated using *different* $w$ cost distributions for different classifiers. But this is inappropriate: the cost distribution must be the *same* for all classifiers we might wish to compare. We would not want to say that, when we used one classifier, we thought that misclassifying a particular object would be likely to incur cost $c_x$, but that, when we used a different classifier, misclassifying that same object would be likely to incur a different cost $c_y$. Putting aside the expense of running an algorithm, we would not want to say that the loss arising from misclassifying a particular object using logistic regression is more severe than the loss arising from misclassifying that object using a support vector machine. Our belief about the likely severities of losses arising from misclassifying objects is an aspect of the problem and researcher, not of the classification method. In general, the distribution we choose for the misclassification cost is independent of what classifier we choose.

Since the AUC is equivalent to (i.e. is a linear transformation of) $L_A$ this also implies that measuring the performance of different classifiers using the AUC is unreasonable, at least if the threshold is chosen to minimise loss for each $c$ (we discuss alternatives in Section 9): it is equivalent to using different measuring criteria for different classifiers, contravening the basic principle of measurement that different objects should be measured using the same or equivalent instruments. This fundamental unsuitability of the AUC as a measure of classifier performance had been previously noted by Hilden (1991), though we were unaware of his work when we wrote Hand (2009). The dependence of the AUC on the classifier itself means that it can lead to seriously misleading results, so it is concerning that it continues to be widely used (see Hand and Anagnostopoulos, 2013, for evidence on how widely used it is).

To overcome the problem, we need to use the same $w(c)$ distribution for all classifiers being compared. This is exactly what the *H-measure* does, and we discuss the choice of *w* in the next section.

As a final point, note that many measures of performance, including the AUC, the proportion correctly classified, and the F-measure, take values between 0 and 1, with larger values indicating better performance. To produce a measure which accords with this convention (Hand, 2009), the H-measure is a standardised version of the loss:
$$H = 1 - L/L_{ref} \qquad [5]$$
where
$$L_{ref} = \pi_0 \int_0^{\pi_1} cw(c)dc + \pi_1 \int_{\pi_1}^1 (1-c)w(c)dc \,, \qquad [6]$$
the value of *L* when $F_0 \equiv F_1$. This reference value of *L* is derived by noting that when $F_0 \equiv F_1$ the minimum of *L* is achieved by classifying everything to class 0 if $c\pi_0 < (1-c)\pi_1$ and everything to class 1 otherwise. That is, classify all objects to class 0 whenever $c < \pi_1$ and to class 1 whenever $c \geq \pi_1$. $L_{ref}$ is the worst case in the sense that it is the minimum loss corresponding to a classifier which fails to separate the score distributions of the two classes at all. Of course, in fact even worse cases can arise – when the classifier assigns classes the wrong way round. For example, the very worst loss is when the classifier assigns all class 0 objects to class 1 and all class 1 objects to class 0, yielding loss $c\pi_0 + (1-c)\pi_1 = \pi_0(2c-1) + 1 - c$. In such cases, we can simply invert all the assigned class labels, yielding a classification performance better than that of random assignment.

## 3. What is a good choice for the weight function?

As described in Hand and Anagnostopoulos (2014), ideally each researcher should contemplate each classification problem and arrive at their own distribution for *c* which best matches their understanding of the problem and its implications. However, while individual problem-specific choice of *w* is the ideal, it is also useful to have a standard conventional *w* that every researcher can use, so that they would produce the same summary measures from the same data. So what properties might a good default choice of *w* have?

Hand (2009) suggested it was reasonable that *w* should decay to 0 as *c* approaches 0 and 1. After all, for example, setting $w(0) \neq 0$ means we believe it is possible that misclassifications of class 0 points incur no cost at all, while misclassifications of class 1 points incur a cost of 1. Under such circumstances it is hardly a classification problem at all, since then the optimum strategy is simply to take every object as class 0. This led to the suggestion that a beta distribution should be used:
$$b(c; \alpha, \beta) = c^{(\alpha-1)}(1-c)^{(\beta-1)} / \int_0^1 c^{(\alpha-1)}(1-c)^{(\beta-1)} dc.$$

Hand and Anagnostopoulos (2014) noted that in dramatically unbalanced applications, such as fraud detection in retail banking, one would typically regard misclassifications of the smaller class (in this example, fraudulent transactions) as more costly than misclassifications of the larger class (legitimate transactions). After all, if a small number of cases (the members of the smaller class) each incurred a small misclassification cost, then very little loss would be incurred by misclassifying them all. This led them to propose that a reasonable choice for the mode of the threshold distribution would be $c = \pi_1$. If class 0 is the larger class, then, since $c$ is the cost of misclassifying objects from class 0, this choice means that misclassifications of this larger class each incur a smaller cost ($\pi_1 = 1 - \pi_0$), whereas if class 0 is the smaller class, then misclassifications of this smaller class each incur a greater cost.

Since the beta distribution has its mode at $(\alpha - 1)/(\alpha + \beta - 2)$, setting this to $\pi_1$ determines one of the two degrees of freedom in the distribution. Hand and Anagnostopoulos (2014) discuss various choices to fix the other and recommend using $\alpha = 1 + \pi_1$, $\beta = 1 + \pi_0$. Thus the basic H-measure uses

$$w(c) = b(c; 1 + \pi_1, 1 + \pi_0).  \quad [7]$$

In summary, the H-measure is given by equation [5] using [3], [6], and [7].

## 4. What does the H-measure mean?

For performance measures to be adopted and used, they must have meaningful and preferably straightforward interpretations. As far as the AUC is concerned, we see from its basic definition, $\int_0^1 F_0(c) f_1(c) dc$, that it is the expected proportion of class 0 objects correctly classified when the expectation is taken over the distribution of the probability scores of class 1 objects. At first glance, this appears to treat the classes asymmetrically, but this is easily remedied. For example, we saw in Hand and Anagnostopoulos (2013) that

$$AUC = \frac{1}{2\pi_0 \pi_1} \int [\pi_0 F_0(c) + \pi_1(1 - F_1(c))] f(c) dc - \frac{\pi_0^2 + \pi_1^2}{4\pi_0 \pi_1}$$

That is, the AUC is a linear function of the expected proportion correctly classified when expectation is taken over the overall mixture distribution of probability scores. Using our earlier terminology, we say that the AUC is "equivalent to" the expected proportion correctly classified when this averaging distribution is used. A similar expression for the AUC in terms of the expected proportion *in*correctly classified can easily be derived.

These straightforward and accessible interpretations have probably helped the widespread adoption of the AUC as a measure of classifier performance. They do not, however, do anything about the fundamental incoherence at its core: in all three of them, the expectation is taken over a distribution which is classifier-dependent, so that the measuring instrument changes according to what is being measured. The fact that the AUC is easy to interpret does not overcome the fact that it is potentially misleading.

The invariance of the AUC to monotonic increasing transformations of the score scale – immediately obvious from its definition – shows that it depends on only the rank order of the test set points, not on their actual numerical scores. AUC can thus be regarded as a measure of *separability* between classes, or refinement. Indeed, for uses other than measuring classification accuracy this is a perfectly valid and useful interpretation. However, as we show in Section 9, this interpretation makes an assumption about the relative importance of the different ranks when combined into an overall measure. This assumption is unlikely to be appropriate for most real practical classification problems.

Another interpretation of the AUC is also often quoted, which also follows immediately from the basic definition: the AUC is the probability that a randomly drawn class 0 object will have a lower score than a randomly drawn class 1 object. This interpretation is fine for some uses of the AUC, but is hardly relevant to most classification problems because, generally, the objects do not arrive in pairs, with one known to be from each class. More typically they arrive individually, or if in groups then we normally do not know how many belong to each class. (There are unusual exceptions.)

But what about interpreting the H-measure? $L_{ref}$ is the expected minimum loss due to misclassification when the classes are not at all separable, that is when the probability score distributions $f_0(s)$ and $f_1(s)$ are identical. This means that $L_{ref} - L$ is the amount that a classifier reduces the expected minimum loss from a classifier which failed to separate the classes at all. The H-measure $H = (L_{ref} - L)/L_{ref}$ is thus *the fractional improvement in expected minimum loss that the classifier yields, compared with a classifier which randomly assigns the points to classes.* Expectation here is of course over whatever *w* distribution the researcher adopts – the beta distribution defined above if our default convention is used.

**5. Is the H-measure a strictly proper scoring rule?**

A "probability scoring rule" (Winkler and Jose, 2010) provides a numerical measure of the accuracy of methods of estimating probabilities. A "proper scoring rule" (Dawid, 2007) is such a measure which has the property that its minimum value occurs when the estimated probability of class membership, *q*, equals the expected probability of class membership; that is, the minimum value occurs when $q = E_y(y|q)$, where $y \in \{0,1\}$ is the true class label. Thus a proper scoring rule takes its minimum value when the probability estimates are calibrated. If this minimum is unique, then the scoring rule is said to be "strictly proper" (Gneiting and Raftery, 2007).

As can be seen from the earlier discussion, a classification performance measure can be described as an average of probability scores (of some kind), averaged over the distribution of predictor vectors describing the objects. Equivalently, classification performance measures are averages of probability scores over the distribution of classification scores resulting from the classification rule. As such it is not really meaningful to ask whether a *classification* performance measure itself is strictly proper or not, but it *is* meaningful to ask whether the probability scoring rule on which it is based is strictly proper.

For the H-measure, the loss for an object with true class label $y \in \{0,1\}$ and classifier-estimated probability *q* of belonging to class 1 is

$$L(q,y) = \int_0^1 \left[c 1_{[q>c]}(1-y) + (1-c)1_{[q \leq c]} y\right] w(c) dc, \qquad [8]$$

where $1_{[a]}$ is the indicator function, taking the value 1 when condition *a* is satisfied and 0 otherwise. Defining $\eta \triangleq E_y(y|q)$, this is a proper scoring rule if

$$argmin_{q \in [0,1]} E_y\bigl(L(q,y)\bigr) = \eta.$$

To see if this is the case, note that

$$E_y(L(q,y)) = \int_0^1 [c1_{[q>c]}(1-\eta) + (1-c)1_{[q\leq c]}\eta]w(c)dc. \qquad [9]$$

$$= (1-\eta)\int_0^q cw(c)dc + \eta \int_q^1 (1-c)w(c)dc. \qquad [10]$$

Differentiating this with respect to $q$ gives

$$dE_y(L(q,y))/dq = [(1-\eta)qw(q) - \eta(1-q)w(q)] = (q-\eta)w(q)$$

This is zero when $q = \eta$. The second derivative at $q = \eta$ reduces to $w(q)$ which is positive on the region of support of $w$, so that $L$ takes its minimum at $q = \eta$, and the loss measure is strictly proper. Note that this does not depend on the shape of $w$.

In Section 9 we review the elegant generalisation by Buja *et al* (2005) which approaches things from the opposite direction.

**6. What about the class sizes?**

For simplicity of exposition, so that we could focus on one source of uncertainty at a time, throughout this discussion we have assumed that the class sizes, $\pi_0$ and $\pi_1$, were known. Uncertainty about the class sizes is usually of a different character from uncertainty about the costs. Class size is an empirical issue, at least in principle subject to exploration by data collection. In particular, the way the training set was drawn will shed light on the underlying population proportions of each of the classes. If the training set was a random sample from the overall population, then the class sizes in the training set can be taken to be proportional to estimates of the class sizes in the population. More generally, biased sampling procedures can be adjusted to yield population class size estimates.

If there is considerable uncertainty about the class sizes (for example, when the future application population is unknown or is expected to evolve over time) then, just as we could choose a distribution for $c$, so we can choose a distribution for $\pi_0$ (and hence by implication also of $\pi_1 = 1 - \pi_0$). Also, although as with the cost distribution this will ideally come from consideration of the problem, a default conventional form can also be useful. Once again, we suggest a beta distribution, this time with mode at ½: $v(\pi_0) = b(\pi_0; 2,2)$.

This leads to

$$H = 1 - \int_0^1 \frac{L(\pi_0)}{L_{ref}(\pi_0)} v(\pi_0) d\pi_0 \qquad [11]$$

where

$$L(\pi_0) = \int_0^1 [c\pi_0(1-F_0(c)) + (1-c)(1-\pi_0)F_1(c)]w(c)dc \qquad [12]$$

$$L_{ref}(\pi_0) = \pi_0 \int_0^{\pi_1} cw(c)dc + \pi_1 \int_{\pi_1}^1 (1-c)w(c)dc, \qquad [13]$$

$$v(\pi_0) = b(\pi_0; 2,2) = 6\pi_0(1-\pi_0) \qquad [14]$$

and

$$w(c) = w(c|\pi_0) = b(c; 2-\pi_0, 1+\pi_0) = \frac{c^{1-\pi_0}(1-c)^{\pi_0}}{\int_0^1 u^{1-\pi_0}(1-u)^{\pi_0} du} \qquad [15]$$

since $1 + \pi_1 = 2 - \pi_0$ in the beta parameters, and where we have explicitly drawn attention to the fact that the distribution of $c$ is conditional on the value of $\pi_0$.

## 7. Estimation

Since the beta distribution $w$ in [15] has the form $c^a(1-c)^b$, the arguments of the integrals in $L_{ref}$ in [13] can also be expressed in terms of beta distributions, yielding

$$L_{ref} = \pi_0 B(\pi_1; 2+\pi_1, 1+\pi_0) \frac{B(2+\pi_1, 1+\pi_0)}{B(1+\pi_1, 1+\pi_0)}$$
$$+ \pi_1 \left[ 1 - B(\pi_1; 1+\pi_1, 2+\pi_0) \frac{B(1+\pi_1, 2+\pi_0)}{B(1+\pi_1, 1+\pi_0)} \right]$$

where $B(c; a, b)$ is the cumulative beta distribution and $B(a,b) = \int_0^1 B(c; a,b)dc$. This is straightforward to evaluate using standard statistical software.

From equation [12], $L(\pi_0)$ is the expected value of $[c\pi_0(1-F_0(c)) + (1-c)(1-\pi_0)F_1(c)]$ when $c$ has distribution $w(c|\pi_0)$, and from equation [11] $H$ is 1 minus the expected value of the ratio $L(\pi_0)/L_{ref}(\pi_0)$ when $\pi_0$ has distribution $v(\pi_0)$. The simplest approach to estimation in terms of coding effort is to estimate these two expected values using a Monte Carlo approach, generating random observations from $w(c|\pi_0)$ and $v(\pi_0)$. An estimate of $F_0(c)$ is given directly by the proportion of class 0 data values which are less than $c$, and an estimate of $F_1(c)$ is given directly by the proportion of class 1 data values which are less than $c$.

## 8. The H-measure from the opposite direction

The discussion above began with a comparison of a probability estimate with a classification threshold. Misclassifications – when class 1 objects had a class 1 estimated probability below the threshold, or class 0 objects had a class 1 estimated probability above the threshold – were taken to incur costs. Balancing these costs over a distribution of likely values for the misclassification cost led to an overall measure of the loss incurred by a classifier.

In contrast, motivated by the need to gain a deeper theoretical understanding of why boosting (Friedman *et al,* 2000) is effective in classification problems, Buja *et al* (2005) begin by focusing on the accuracy of the probability estimates themselves, writing a general loss for class 1 estimated probability $q$ and true class $y$ as

$$L(q, y) = (1-y)L_0(q) + yL_1(1-q),$$

where $L_0(q)$ is the loss due to misclassifying a class 0 point when the estimated class 1 probability is $q$, with similar for $L_1(q)$. They then used a result described in Shuford *et al* (1966), Savage (1971), Dawid (1986), and Schervish (1989) which shows that if $L_0$ and $L_1$ are differentiable then $L$ is a proper scoring rule if and only if $L'_0(q) = w(q)q$ and $L'_1(1-q) = w(q)(1-q)$ for some weight function $w(q) \geq 0$ on (0,1) satisfying certain conditions. Moreover the scoring rule is strictly proper if $w(q) > 0$ almost everywhere on (0,1).

This leads to

$$E_y(L(q, y)) = (1-\eta)L_0(q) + \eta L_1(1-q).$$
$$= \int_0^1 [(1-\eta)c 1_{[q>c]} + \eta(1-c)1_{[q\leq c]}]w(c)dc$$

$$= (1-\eta)\int_0^q cw(c)dc + \eta\int_q^1 (1-c)w(c)dc,$$

which is equation [10]. That is, if the probability scoring rule is proper then it has the form of a mixture of cost-weighted misclassification losses, with a distribution $w(c)$ for the costs. Thus Buja *et al* (2005) have proceeded in the opposite direction from us. Where we began with the cost-weighting and ended with the rule being proper, their result shows that the cost-weighted representation is natural, and not merely an artefact of a way of describing the classification problem.

Buja *et al* (2005) explore possible choices for *w*, showing how appropriate choices lead to log-loss (the negative of the log-likelihood of the Bernoulli model), squared error loss, and boosting loss. Like us (Hand and Anagnostopoulos, 2014), they also note the merits of using a Beta distribution as *w*. In our definition of the H-measure, we have taken $w(c) = b(c; 1+\pi_1, 1+\pi_0)$, as in [7].

In summary, while we began our derivation of the H-measure from the recognition that the AUC used different cost-weighting distributions for different classifiers, Buja *et al* (2005) have shown that the use of different cost-weighting distributions is equivalent to using different loss functions when evaluating the estimated probability of class 1 membership: general versions of the H-measure, with carefully chosen *w* distributions, are the essence of measuring classifier performance. And from the perspective of the route we took, the work of Buja *et al* (2005) demonstrates that the AUC is equivalent to using different probability scoring functions to evaluate the accuracy of the estimates of the probability of belonging to class 1 for different classifiers.

## 9. Other ways of choosing the classification threshold

In Section 2 we said that, for any given cost *c*, a "sensible" choice of classification threshold *t* is that which minimises $L(c;t)$. However, other ways of choosing *t* are possible. In an extensive and comprehensive investigation, Hernández-Orallo *et al* (2012) term this dimension of classifier evaluation the *threshold choice method*, exploring various ways the threshold might be chosen. They cross-classify this dimension with the *operating condition* dimension of classifier evaluation, which includes such things as the misclassification cost distribution and class imbalance, to produce a unified way of looking at a number of classifier performance metrics.

One special case arises when the classification threshold is chosen independently of the cost *c*. We see from [1] that, when the cost is chosen from a distribution $w(c)$ and the threshold independently chosen from a distribution $u(t)$ we obtain the overall expected loss

$$L = \int_0^1 \int_0^1 [c\pi_0(1-F_0(t)) + (1-c)\pi_1 F_1(t)]w(c)u(t)dc\,dt$$
$$= \int_0^1 [E(c)\pi_0(1-F_0(t)) + (1-E(c))\pi_1 F_1(t)]u(t)dt \qquad [16]$$

It is clear from [16] that this leads to an *L* which is equivalent to taking all misclassifications of a class 0 object as incurring cost $E(c)$ and all contrary misclassifications as incurring cost $1-E(c)$, regardless of the classification threshold. If this approach were to be adopted, it then seems strange adopting a distribution for *t* at all, since *L* can be minimised by choosing $u(t)$ to be a delta function located at the threshold *t* which satisfies
$$E(c) = \pi_1 f_1(t)/f(t)$$
as we saw in [2]. If, nonetheless, one does wish to adopt a distribution for *t*, then some rationale has to be found for making the choice of its form. Flach *et al* (2011) have explored this, taking a uniform distribution for *c*, so that $E(c) = 1/2$, and adopting $u(t) = f(t)$. From [16] we see immediately that this leads to a linear function of the AUC. On this basis, Flach *et al* (2011) claim to have derived a coherent interpretation of the AUC. This is correct, but, as they also point out, taking $u(t) = f(t)$ is equivalent to choosing, with equal probability, the scores of each of the test objects to act as the threshold. That is, it is equivalent to using only the rank-order information in the scores of the test set objects.

Now, by definition, choosing *t* on the basis of *c* (for example, to minimise the loss for each possible value of *c* or to minimise the loss for the expected value of *c*) requires relating the *t* scale to the *c* scale, and this presupposes a calibration of the scores to some numerical scale – often, but not necessarily, a probability scale. However, if the test set objects are simply *rank-ordered*, without an associated numerical scale, then *t* cannot be chosen in this way and the choice must be based on choosing a rank as the threshold (or, more formally, just below some rank). For example, we could say we will classify as class 0 those test set objects which rank in the lowest 10%, or perhaps the lowest 90%, of all test set objects and so on. Then we could evaluate performance by seeing what proportion of the class 0 objects have been correctly classified as class 0. And there are variants of this. For example, we could set the threshold to select the lowest 10% ranks of class 0 test set objects (instead of *all* test set objects), and see what overall misclassification rate this led to. These sorts of approaches correspond to variants of *screening*, with the proportion to be selected fixed in advance. They are useful strategies provided one can decide beforehand what is an appropriate proportion to choose.

Unfortunately, in many situations it is difficult to know what proportion (or rank) one might want to choose – just as it was difficult to choose costs, as discussed above. In such cases, an alternative to picking a particular proportion or rank to define the threshold is to average the performance (the proportion of class 0 objects correctly classified, the misclassification rate, or whatever you like) if we were to choose the proportion according to some distribution. For example, we could with *equal probability* choose the rank of each of the class 1 test set points as the thresholds. If we then take, as the performance measure, the average proportion of class 0 points correctly classified using these thresholds, we obtain the AUC. This is what Flach *et al* (2011) have done.

Choosing the threshold uniformly from the ranks of the class 1 objects immediately prompts the question of why accord them equal probability of being chosen? We could have put more weight on the lower ranks, for example. To illustrate, I might know that only a small proportion of the population suffer from a particular disease, so I will want my classifier to flag up only a small proportion of the incoming people, even if I do not know exactly what small proportion I will be able to investigate more rigorously. (If I'm trying to identify people who are likely to be suffering from Stoneman Syndrome – about 1 in 2 million people – I will not want to flag 90% of the population as possibly having it and deserving more detailed investigation).

One answer might be that the uniform distribution represents ignorance about the circumstances under which the classifier will be applied in the future. But that's not really right: it represents a particular choice. For example, if $n_1$ is the number of class 1 test set objects, the uniform distribution says that we think it equally likely that we will want to classify a proportion $1/n_1$ and a proportion $(n_1 - 1)/n_1$ of the class 1 points as belonging to class 1. That's a very strong assumption (and, worse, it's hidden: people are making this assumption without realising it). To the extent that it is a poor representation of the circumstances under which the classifiers will be applied, it could give a very misleading impression of relative performance.

If the scores *have been* calibrated to some numerical scale, the question of "why choose a uniform distribution on the ranks?" corresponds to "why do we believe that threshold values occurring near the mode of the estimated class 1 probability distribution for the particular classifier, $f(t)$, should be preferable to other values?". This leads us back to the fact that $f(t)$ is classifier-dependent, and the inappropriateness of the AUC for comparing classifiers if numerical scores have been produced (such as estimated class 1 membership probabilities).

**10. Conclusion**

Hand and Till (2001) describe the widely used AUC measure of classification performance as concentrating "attention on how well the rule differentiates between the distributions of the two classes, … not influenced by external factors which depend on the use to which the classification is to be put." It is an "internal" measure, making no reference to matters such as the relative severity of different types of misclassification. However, the assessment of a classification measure cannot take place in a vacuum, but must always depend on the application: in assessing a classifier we are asking "how good is the classifier for such and such a role?", not least because a classifier good in one application may be poor in another. This is why Hand (1997) concludes by saying "The opening sentence of this book posed the question, Which is the best type of classification rule? We can now answer that question. The answer is: It depends …". One consequence of the failure to take context into account is that the AUC is incoherent, in the sense that it is equivalent to the expected proportion of class 0 objects misclassified (or class 1 objects, or overall objects, or correctly classified) but where the distribution yielding the expected value varies from classifier to classifier. It is not comparing like with like. The AUC is certainly a coherent measure of *separability* between classes, since it is based solely on the order of scores of the test set objects, and not on the numerical values of those scores. However, separability is not the same as classification performance since separability appears to ignore the central role of the classification threshold. But this is deceptive: implicit in the AUC when viewed in this way is a hidden assumption that each rank of the test set objects is equally likely to be chosen as the threshold. This is an unrealistic assumption in most practical applications.

The H-measure tackles the incoherence of the AUC by introducing costs for the different types of misclassification. If the costs were known, the optimal threshold would be known and the problem would reduce to that of summarising a standard two-by-two confusion matrix of true class labels by predicted class labels. However, it is common that precise values of the costs are not known at the time that the classifiers must be evaluated – earlier we gave the example of evaluating a medical diagnostic system when the future misclassification severity would depend on what treatment options would become available. To tackle such situations, we take the expectation over a distribution of likely cost values. Ideally this distribution should be chosen by the researcher on the basis of their knowledge of the problem – like a Bayesian prior. And, indeed, we recommend that researchers always try to come up with such a distribution. However, it is also useful to have a standard default distribution which can be used to yield a conventional measure. We have recommended a particular form of beta distribution for this role. We then generalised to cover the case when the class sizes are also unknown.

More generally, Buja *et al* (2005) have demonstrated that choice of different functions for the cost distribution is equivalent to using different loss functions for different classifiers when estimating the class membership probabilities. It seems difficult to see why one would want to minimise (for example) log-loss when using a neural network, but squared error loss when using a random forest.

Since the introduction of the H-measure in 2009, and its subsequent refinement in later years, researchers have raised various questions about it, such as how to interpret it and whether it is strictly proper. This paper has attempted to provide answers to these questions.